\documentclass[runningheads]{llncs}

\usepackage{lipsum}
\usepackage{xcolor}
\usepackage{multirow}
\usepackage{booktabs}
\usepackage{adjustbox}
\usepackage[para]{threeparttable}
\usepackage{graphicx}
\usepackage[hidelinks]{hyperref}
\usepackage{floatrow}
\usepackage{mathptmx} 
\usepackage{times} 
\usepackage{amsmath} 
\usepackage{amssymb}  
\usepackage{bbm}

\usepackage{xfrac}

\usepackage{amsmath}

\DeclareMathOperator*{\argmin}{arg\,min}

\usepackage{pifont}
\newcommand{\xmark}{\ding{55}}%

\usepackage{caption}
\usepackage{subcaption}
\usepackage{orcidlink}

\usepackage{csquotes}

\begin{document}

\title{Reading Between the Frames: Multi-Modal Depression Detection in Videos from Non-Verbal Cues}

\titlerunning{Multi-Modal Depression Detection in Videos from Non-Verbal Cues}

\author{
David Gimeno-Gómez\inst{1}$^\dagger$\orcidlink{0000-0002-7375-9515} \and
Ana-Maria Bucur\inst{1,2}$^\dagger$\orcidlink{0000-0003-2433-8877} \and Adrian Cosma\inst{3}$^\dagger$\orcidlink{0000-0003-0307-2520} \and \\Carlos-David Martínez-Hinarejos\inst{1}$^\ddagger$\orcidlink{0000-0002-6139-2891} \and Paolo Rosso\inst{1,4}$^\ddagger$\orcidlink{0000-0002-8922-1242}}

\authorrunning{D. Gimeno-Gómez et al.}

\institute{PRHLT Research Center, Universitat Politècnica de València, Spain\\
\email{dagigo1@dsic.upv.es,cmartine@dsic.upv.es}\\
\and Interdisciplinary School of Doctoral Studies, University of Bucharest, Romania\\ \email{ana-maria.bucur@drd.unibuc.ro} \and Politehnica University of Bucharest, Romania\\ \email{ioan\_adrian.cosma@upb.ro} \and ValgrAI Valencian Graduate School and Research Network of Artificial Intelligence, Spain\\
\email{prosso@dsic.upv.es}}

\maketitle              

\def\thefootnote{$\dagger$}\footnotetext{Equal contribution.}\def\thefootnote{\arabic{footnote}}
\def\thefootnote{$\ddagger$}\footnotetext{Equal supervision.}\def\thefootnote{\arabic{footnote}}

\begin{abstract}
Depression, a prominent contributor to global disability, affects a substantial portion of the population. Efforts to detect depression from social media texts have been prevalent, yet only a few works explored depression detection from user-generated video content. In this work, we address this research gap by proposing a simple and flexible multi-modal temporal model capable of discerning non-verbal depression cues from diverse modalities in noisy, real-world videos. We show that, for in-the-wild videos, using additional high-level non-verbal cues is crucial to achieving good performance, and we extracted and processed audio speech embeddings, face emotion embeddings, face, body and hand landmarks, and gaze and blinking information. Through extensive experiments, we show that our model achieves state-of-the-art results on three key benchmark datasets for depression detection from video by a substantial margin. Our code is publicly available on GitHub\footnote{\url{https://github.com/cosmaadrian/multimodal-depression-from-video}}. 
\end{abstract}



\begin{displayquote}
{\small \textit{"There is only continual motion. If I rest, if I think inward, I go mad."}}

\mbox{}\hfill {\small\textit{The Unabridged Journals of Sylvia Plath}}
\end{displayquote}

\section{Introduction}
\label{sec:introduction}
Depression is a leading cause of disability, with 5\% of the adults worldwide suffering from it\footnote{\href{https://www.who.int/health-topics/depression}{https://www.who.int/health-topics/depression}. Accessed August 28th, 2023}. 
According to the World Health Organization (WHO), the prevalence of depression was up to 25\% percent during the first year of the COVID-19 pandemic\footnote{\href{https://www.who.int/news/item/02-03-2022-covid-19-pandemic-triggers-25-increase-in-prevalence-of-anxiety-and-depression-worldwide}{https://www.who.int/news/item/02-03-2022-covid-19-pandemic-triggers-25-increase-in-prevalence-of-anxiety-and-depression-worldwide}. Accessed August 28th, 2023}. 
Many efforts have been directed at detecting depression from text on social media sites such as Reddit \cite{anxo2023psyprof,trifan2020understanding,rissola2020beyond} and Twitter \cite{shen2017depression,benton-etal-2017-multitask,leis2019detecting}. This can be attributed to the relative ease of annotation (self-mentions \cite{yates-etal-2017-depression,zanwar-etal-2023-smhd}, certain subreddits participation \cite{pirina-coltekin-2018-identifying,haque2021deep}) and abundance of available textual data that can be retrieved in a relevant context (i.e., as a thread). 
Although research on multi-modal depression detection from social media data has been performed using textual, visual, and online behavioral data, visual data is mostly limited to images \cite{reece2017instagram,gui2019cooperative,birnbaum2020identifying,FernandezBarrera2022evaluating,bucur2023matter,yadav2023towards}. However, many social media sites (YouTube, TikTok, Instagram) focus on user-generated video content and, currently, depression detection from in-the-wild video remains largely unexplored. The video modality poses an additional set of challenges for depression detection as it is significantly noisier, video length is highly variable (from shorts of a few seconds to half-hour vlogs in which the person talks directly to the camera), the face and body are not always visible, context rapidly changes, and audio quality is low, especially in amateur videos. Uploaded videos are mostly out of context and stand-alone as opposed to interactive discussions in social media threads. In this context, the task is to process subtle depression cues from the subject's behavior present in the video. 

According to the DSM-5 criteria \cite{APA2013}, major depressive disorder (depression) is manifested through symptoms such as depressed mood, sleep disturbance, appetite changes, loss of interest, etc. In addition, psychomotor changes (agitation or retardation) are a central feature of depression, intertwined with other symptoms, such as loss of energy, fatigue, and lack of concentration \cite{bennabi2013psychomotor}. Some of the psychomotor particularities associated with depression are: reduced facial expressiveness \cite{renneberg2005facial}, slower speech rate, longer pause time \cite{yamamoto2020using}, and downward head tilt \cite{fiquer2013talking}. In addition, studying psychomotor changes in depression is essential because they can predict clinical response to medication in depression \cite{taylor2006psychomotor,yamamoto2020using}. However, these cues have mainly been studied in laboratory settings \cite{gratch2014distress,ringeval2019avec} and significantly less so in in-the-wild videos. 
To the best of our knowledge, the only available benchmark dataset for depression detection using data from online platforms (i.e., YouTube) is the D-Vlog dataset \cite{yoon2022dvlog}.  

In this work, we propose a multi-modal temporal model that processes multiple non-verbal cues across time to estimate depression from both noisy, in-the-wild, videos \cite{yoon2022dvlog} and from controlled, laboratory recordings \cite{gratch2014distress,ringeval2019avec}. Our method is simple and flexible: by using appropriate high-level modality extractors, positional embeddings and modality conditioning vectors, this approach can be easily scaled to an arbitrary number of modalities. 
We surpass state of the art on three datasets: two controlled datasets (DAIC-WOZ \cite{gratch2014distress} and E-DAIC \cite{ringeval2019avec}) and an in-the-wild dataset (D-Vlog \cite{yoon2022dvlog}). Through extensive experiments, we show that using the appropriate data modalities and semantic embeddings is crucial in processing non-verbal cues from a relatively small amount of videos. This work makes the following contributions:

\begin{enumerate}
    \item We propose a simple and flexible multi-modal architecture that can process non-verbal depression cues from an arbitrary number of modalities across time. Our model achieves state-of-the-art results on three key depression detection datasets, obtaining 0.78 F$_{1}$ on D-Vlog \cite{yoon2022dvlog}, 0.67 F$_{1}$ on DAIC-WOZ \cite{gratch2014distress}, and 0.56 F$_{1}$ on E-DAIC \cite{ringeval2019avec}, surpassing the previous state of the art \cite{yoon2022dvlog,nguyenmultimodal,zheng2023two,zhou2022tamfn,zhou2023caiinet,tao2023depressive,song2018daicwoz,ma2016daicwoz,wei2023daicwoz,williamson2016daicwoz}.

    \item We show that, for in-the-wild scenarios (e.g., for D-Vlog \cite{yoon2022dvlog}), using appropriate high-level semantic embeddings is crucial, and we explore additional non-verbal cues informed by studies in psychology \cite{bennabi2013psychomotor,renneberg2005facial,fiquer2013talking,yamamoto2020using}: emotion-informed face embeddings, task-agnostic audio embeddings, body and hand landmarks, and eye movements (blinking and gaze). Using the additional modalities, our model exhibits a markedly better performance in in-the-wild depression detection.
    
    \item We show that our model is interpretable by using Integrated Gradients \cite{sundararajan2017axiomatic}, estimating the relevance of each modality across time for a particular subject. In this way, our method is a potentially valuable tool in preventive screening for psychologists.
\end{enumerate}

\section{Related Work}
\label{sec:related-work}
\noindent \textbf{Depression detection from video.} Depression detection from video content \cite{he2022deep} follows the same high-level pipeline from general multi-modal video classification \cite{pampouchidou2017automatic}, which has been fueled by advances in multi-modal classification in images \cite{sleeman2022multimodal,alayrac2020self,jaegle2021perceiver}. 
Similar to general multi-modal video classification, depression detection from video presumes the (automatic) extraction of temporal and spatial features and processing them with a classifier \cite{pampouchidou2017automatic}. Some low-level features, extracted by using classical image and audio processing algorithms, are: loudness, spectral flux, phoneme duration, pitch slopes for audio and facial landmarks, facial action units \cite{friesen1978facial}, and histogram of oriented gradients features for video content \cite{pampouchidou2016depression,williamson2014vocal}. High-level semantic features include facial expressions, smile intensity and duration, head movement, etc. \cite{pampouchidou2017automatic}. Most current methods proposed for depression detection \cite{tao2023depressive,zhou2023caiinet,zhou2022tamfn,zheng2023two} use facial landmarks and low-level acoustic descriptors \cite{yoon2022dvlog}, disregarding the temporal information \cite{nguyenmultimodal,zhou2023caiinet,yoon2022dvlog}. More recently, multi-modal large-language models (LLMs) offer a promising paradigm for video classification, with approaches such as MiniGPT-4 \cite{anonymous2023minigpt} and PandaGPT \cite{su2023pandagpt}. In particular, some models are oriented towards mental health prediction from online text \cite{xu2023leveraging} with approaches such as MentalLLaMa \cite{yang2023mentalllama}. However, in this work, we focus on processing a plethora of non-verbal cues, disregarding raw pixel data and transcripts, which current LLMs are not suitable.

\noindent \textbf{Benchmarking datasets.} Benchmark datasets for studying the non-verbal behavior of subjects with depression include mostly data collected in laboratory settings, e.g., the DAIC-WOZ  \cite{gratch2014distress}, E-DAIC \cite{ringeval2019avec}, Depression Severity Interviews Database \cite{dibekliouglu2017dynamic}, Audio-Visual Depressive Language Corpus \cite{valstar2013avec}, and the BlackDog Database \cite{alghowinem2012joyous}. These datasets consist of clinical interviews or footage with subjects performing different tasks (reading certain texts, counting, etc.). However, research in depression detection from video in general, in-the-wild, scenarios is limited. To the best of our knowledge, D-Vlog \cite{yoon2022dvlog} is the only dataset containing in-the-wild user-generated videos.

\noindent \textbf{Present work.} Inspired by psychological research \cite{renneberg2005facial,fiquer2013talking,rottenberg2008emotion,mackintosh1983blink,lucas2015towards}, we employ additional features related to motor manifestations of depression, such as face and audio embeddings, hand and body landmarks, blinking, and gaze patterns. Our relatively simple architecture is able to process multiple modalities across different frame rates and processes only relevant parts of the video to perform classification. In contrast to previous approaches that mainly use global information \cite{nguyenmultimodal,zhou2023caiinet,yoon2022dvlog}, we design our architecture to handle both local, frame-level information, as well as temporal video dynamics.

\section{Method}
\label{sec:method}
\begin{figure}[hbt!]
    \centering
    \includegraphics[width=1.0\linewidth]{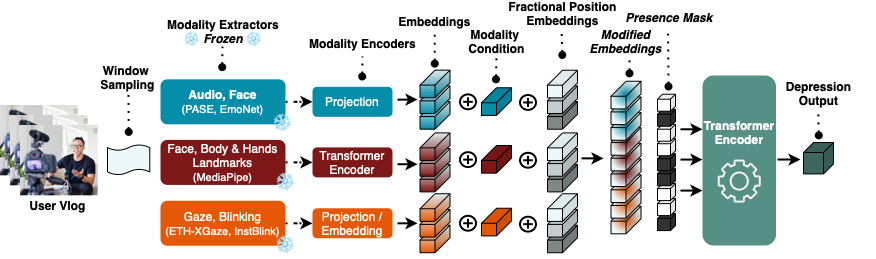}
    \caption{The overall architecture of our proposed method. We extract high-level non-verbal cues using pretrained models, process them using a modality-specific encoder, condition the resulting embeddings with positional and modality embeddings, and process the sequence with a transformer encoder to perform the final classification.
    }
    \label{fig:architecture-diagram}
\end{figure}

\subsection{Model Architecture}

\noindent \textbf{Overview.} Given a dataset of vlogs with people directly talking to the camera, our aim is to train a model to estimate whether the subject in a video has depression or not. In this formulation, depression detection amounts to video-level classification. However, performing the classification of a high-level psychological affliction directly from pixels is infeasible without impractical amounts of labeled data. Consequently, in our work, we considered only high-level non-verbal cues, and discarded any textual information to remove any conversational topic bias \cite{wolohan-etal-2018-detecting}. The way most depression datasets are collected is based on direct mention of diagnosis \cite{yates-etal-2017-depression,zanwar-etal-2023-smhd} and by using only non-verbal cues, the model's performance is a more accurate depiction of its performance in realistic settings. Non-verbal cues are extracted using state-of-the-art pretrained models \cite{ravanelli20pase,toisoul2021emonet,bulat2017far,hongyi2020ghum,zhang2020ethxgaze,zeng2023blink} and subsequently processed to perform classification. Figure \ref{fig:architecture-diagram} provides a high-level overview of our pipeline for depression classification from video. We further present a way to properly handle videos of different framerates, video lengths, and process an arbitrary amount of modalities.

Formally, given a dataset of depression labeled videos $\mathcal{D} = \{(V_1, \hat{y}_1), \dots (V_n, \hat{y}_n)\}$, where $V_i$ is the video and $\hat{y}_i$ its corresponding label, the goal is to find the optimal parameters $\theta$ of a model $f_{\theta}$ that minimize the average cross entropy loss across the dataset: $ \hat{\theta} = \argmin_\theta \mathbb{E}_{1 \leq k \leq n}[\mathcal{L}(f_\theta(V_k), \hat{y}_k)]$.


\noindent \textbf{Window Sampling.} Since videos and vlogs in the wild have vastly different durations and framerates (see Fig \ref{fig:video-duration}), with some videos exceeding 30 minutes, it is computationally unfeasible to process them directly in their entirety. Consequently, we operate on randomly sampled temporal windows $W_l^k \sim V_k$ from each video, with $W_l^k$ of fixed time in seconds, and $l$ a random temporal index in the video. Temporal windows can cover a different number of total frames, corresponding to each video's framerate. The audio is sampled at 100 frames per second \cite{gales2008application}, but video framerates from D-Vlog \cite{yoon2022dvlog} vary between 6 and 30 frames per second. Through the window sampling approach, we assume that the mental health information in the video is time invariant.

\noindent \textbf{Modality Extraction and Encoding.} 
For each video, we considered multiple semantic modalities $\mathcal{M} = \{m_1, m_2 \dots \}$, that are extracted using a frozen pretrained model from both audio and visual information of the video at each frame. Consequently, the total number of frames for a video across all temporally concatenated modalities is the sum $T^k = T^k_1 + \dots + T^k_{|\mathcal{M}|}$. Since each modality has a different dimensionality $d_{m_j}$, we uniformize each modality output with an associated learnable modality encoder $E_{m_j}: \mathbb{R}^{d_{m_j}} \rightarrow \mathbb{R}^{d}$, operating at the frame-level, such that each modality has the same dimensionality $d$. Thus, the feature vector for a window $W_l^k$ and a modality $m_j$ is $x_{m_j} = E_{m_j}(W_l^k)$, with $x_{m_j} \in \mathbb{R}^{T_j \times d}$. Details about each modality encoder $E_{m_j}$ are showcased in Section \ref{sec:non-verbal-cues}. Furthermore, each modality extractor can signal the presence of a modality in a frame, which allows us to construct a binary presence mask $M_j^k \in \mathbb{R}^{T_j^k}$. The presence mask is later used in attention masking for the final transformer encoder \cite{vaswani2017attention}. In practice, for in-the-wild videos, we first extracted and cached all modalities for each video, and sampled windows such that the subject is present in at least 50\% of the frames in a window. This is because some videos were amateur-made and extremely noisy, and the subject's body, hands and face were not always visible; for instance, some people talk while driving their car and the steering wheel obscures the camera, or the subject walks out of the scene.

\noindent \textbf{Modality Condition.} We distinguish between the embeddings of different modalities by using an additional learned condition embedding $e_{m_j} \in \mathbb{R}^{|\mathcal{M}| \times d}$ that is added to each vector of its corresponding modality. This is similar to how BERT-type models \cite{devlin-etal-2019-bert} differentiate between sentences in tasks such as natural language inference. 

\begin{figure}[hbt!]
    \centering
    \includegraphics[width=0.7\textwidth]{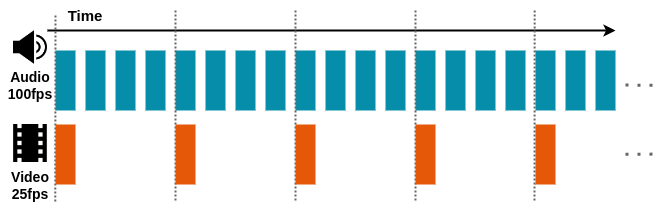}
    \caption{Illustrative example of fractional positional embedding for temporally aligning video and audio sampling rates, similar to the work of Harzig et al. \cite{harzig2022fractional}.}
    \label{fig:fractional-positional-embedding}
\end{figure}

\noindent \textbf{Fractional Position Embeddings.} Due to the mismatch between sampling rates, we temporally align the frames of acoustic and video-based modalities by using fractional positional embeddings \cite{harzig2022fractional} of the form $p_{m_j} \in \mathbb{R}^{\max_i(T_i) \times d}$. For a video, we construct a matrix of positional embeddings of size $\max_i(T_i)$, corresponding to the modality with the highest sample rate, usually audio. Afterwards, positions for a modality $m_j$ are then uniformly indexed according to the ratio $r = \lfloor \max_i(T_i) / T_{m_j} \rfloor$. Fractional positional embeddings are based on sinusoidal positional embeddings \cite{vaswani2017attention} and are not learned during training. Frames across modalities have the same positional embedding at the same corresponding positions in time, irrespective of the sampling rate (see Figure \ref{fig:fractional-positional-embedding}). Readers are referred to Harzig et al. \cite{harzig2022fractional} for more details.

An individual modified feature vector becomes $\bar{x}_{m_j} = x_{m_j} + e_{m_j} + p_{m_j}$. Finally, we process the concatenated modified feature vectors $X_l^k = \bar{x}_{m_1} \mathbin\Vert \bar{x}_{m_2} \mathbin\Vert \dots \mathbin\Vert \bar{x}_{|\mathcal{M}|}$ alongside each corresponding presence mask $M_l^k = M_1 \mathbin\Vert M_2 \mathbin\Vert \dots \mathbin\Vert M_{|\mathcal{M}|}$ from each modality of a sampled temporal window with a transformer encoder network \cite{vaswani2017attention} to obtain a prediction $y = f_{\theta}(X_l^k, M_l^k)$, with $X_l^k \in \mathbb{R}^{T_l^k \times |\mathcal{M}| \times d}$ and $M_k \in  \mathbb{R}^{T_l^k \times |\mathcal{M}|}$. The final optimization formulation can be described as $ \hat{\theta} = \argmin_\theta \mathbb{E}_{(1 \leq k \leq n, 1 \leq l \leq |V_k|)}[\mathcal{L}(f_\theta(X_l^k, M_l^k), \hat{y}_k)]$.




\noindent \textbf{Discussion.}
In our framework, modality fusion is done immediately after encoding each modality with the learnable encoders $E_{m_j}$, and the predominant part of the computation is performed with the main transformer encoder. This type of fusion can be considered ``early" fusion \cite{sleeman2022multimodal}. Furthermore, as defined by Vaswani et al. \cite{vaswani2017attention}, the scaled dot-product attention
is permutation invariant across the sequence without the addition of positional embeddings \cite{wang2020position}. However, we employ two types of positional embeddings: learned modality conditions and fractional positional embeddings. In a single timestep $t$, the modality vectors $x^t_{m_j}$ are permutation invariant, as they share the same positional embedding. The order of the modalities does not affect the final computation, as the modalities are considered an unordered set, corresponding to the SetTransformer \cite{lee2019set} formulation.
Other approaches \cite{nguyenmultimodal,zhou2023caiinet,yoon2022dvlog} mainly perform a global computation across a coarse temporal sampling of modalities, which is prone to lose the information of the finer non-verbal cues in manifestations of depression. However, through our window sampling, modality conditions and positional embeddings, the transformer attention operates both locally, inter-modality (at the same timestep), and globally, intra-modality (across multiple timesteps), to perform a successful depression classification. In this setting, the model performs a soft cross-attention between modalities. Furthermore, compared to other approaches in this area \cite{zheng2023two,zhou2022tamfn,zhou2023caiinet}, which propose overly-complicated methods with modest results, our approach is simple, flexible, easy to reproduce, and it surpasses other methods on the three main datasets we evaluated.


\subsection{Extracting and Encoding Non-Verbal Cues}
\label{sec:non-verbal-cues}
\noindent \textbf{Psychological Motivation.}
Psychomotor manifestations are one of the main features of depression \cite{bennabi2013psychomotor} found in the DSM-V criteria and self-assessment tools, such as Patient Health Questionnaire-9 (PHQ-9) or Beck Depression Inventory-Second Edition (BDI-II). The non-verbal facial cues associated with depression are reduced expressiveness \cite{renneberg2005facial}, and fewer smiles \cite{girard2013social}. In terms of body language, a downward head tilt \cite{fiquer2013talking}, longer gait cycles, and a slower gait cadence \cite{zhang2022associations} are common indicators. Regarding hand movements, individuals with depression may use fewer gestures \cite{rottenberg2008emotion} and exhibit impaired gesture performance \cite{pavlidou2021hand}. A slower speech rate characterizes the voice, with longer pauses between words \cite{yamamoto2020using}, and a decrease in loudness \cite{wang2019acoustic}. The eyes undergo subtle shifts with an increased blinking rate \cite{mackintosh1983blink}, reduced eye contact, and increased gaze angle \cite{lucas2015towards}. Given these non-verbal changes in depressed individuals, we investigated how they interact and complement each other in depression detection. This work refers to major depressive disorder (MDD), not other related mood disorders, such as bipolar disorder, peripartum disorder, etc.

\noindent\textbf{Audio-Visual Embeddings (AV).} We extract semantic embeddings for the audio channels and the face gestures using pretrained models. For \textbf{audio embeddings}, we used the Pyannote Audio toolkit\footnote{\href{https://github.com/pyannote/pyannote-audio}{https://github.com/pyannote/pyannote-audio}. Accessed August 28th, 2023} \cite{bredin2020pyannote,bredin21segmentation} to detect voice activities and extract 256-dimensional voice embeddings using the PASE+ model \cite{pascual19pase,ravanelli20pase} in the time slots in which the person is speaking. The PASE+ model provides general embeddings \cite{pascual19pase,fang2023avemotion} as it was trained in a task-agnostic manner in noisy and reverberant environments. For extracting emotion-informed \textbf{face embeddings}, we used EmoNet\footnote{\href{https://github.com/face-analysis/emonet}{https://github.com/face-analysis/emonet}. Accessed August 28th, 2023} \cite{toisoul2021emonet}, a model capable of estimating both discrete emotion classes and their corresponding valence and arousal \cite{russell1980circumplex,gunes2013affect} measures. We obtain a 256-dimensional embedding vector by taking the embedding from the last layer before the classification of EmoNet. For both types of embeddings, in our model we process the embeddings with a 1D batch normalization layer followed by a linear layer to project the embeddings into a 256-dimensional space.

\noindent\textbf{Landmarks (LM).} We extract relevant facial, body and hands keypoints using pretrained networks. For \textbf{face landmarks}, we used the Face Alignment Network model \cite{bulat2017far} to extract 68 2-dimensional facial landmarks alongside their corresponding confidence score. For \textbf{body landmarks}, we used the MediaPipe\footnote{\href{https://github.com/google/mediapipe}{https://github.com/google/mediapipe}. Accessed August 28th, 2023} \cite{hongyi2020ghum,bazarevsky2020blazepose} toolkit to extract 33 3-dimensional landmarks. For \textbf{hand landmarks}, we used the hand landmark detector from the MediaPipe \cite{hongyi2020ghum} toolkit to extract 21 3-dimensional landmarks for each hand. In our model, we process the landmarks using a 1D batch-normalization layer followed by a 2-layer transformer encoder that operates on the spatial dimension, obtaining the final embedding through average spatial pooling. Additionally, for encoding hand landmarks we distinguish between right and left hands by adding an embedding vector, similar to the modality condition described above. 

\noindent\textbf{Eye Information (EYES).} We extract gaze and blinking information as additional modalities. For \textbf{gaze tracking} we used the official implementation\footnote{\href{https://github.com/hysts/pytorch\_mpiigaze\_demo}{https://github.com/hysts/pytorch\_mpiigaze\_demo}. Accessed August 28th, 2023} of the ETH-XGaze gaze estimator \cite{zhang2020ethxgaze} to extract the gaze direction represented by 3 angle coordinates. For \textbf{blinking patterns}, we used the official implementation of the InstBlink model\footnote{\href{https://github.com/wenzhengzeng/MPEblink}{https://github.com/wenzhengzeng/MPEblink}. Accessed August 28th, 2023} \cite{zeng2023blink}, and labeled the video frames where the person is blinking (0 for not blinking and 1 for blinking). To encode gaze, we process the angles using a 1D batch normalization followed by a linear layer to project the angle vector into a 256-dimensional embedding. To encode blinking, at each frame we used one of two different learnable 256-dimensional embeddings, corresponding to blinking or not blinking states.

\section{Experiments}
\label{sec:experiments}

\subsection{Datasets}
\label{sec:datasets}

\noindent \textbf{D-Vlog} \cite{yoon2022dvlog} is an in-the-wild dataset for multi-modal depression collected from user-generated YouTube vlogs. It comprises 961 video samples (159.2 hours) with an average duration of 9.9 minutes, containing 406 control and 555 depressed subjects. It is split into a training set with 647 samples, a validation set with 102 samples, and a test set with 212 samples. D-Vlog is manually annotated following overt mentions of depression symptoms such as mentions of suicidal thoughts, mentions of depression medication and so on. In our work, in addition to studying the original audio-visual features provided by the authors, we explored the multiple non-verbal cues described in Subsection \ref{sec:non-verbal-cues}. When extracting additional features, some of the videos were unavailable (deleted or made private by the owner), resulting in 861 videos. Specifically, 70, 8, and 23 videos were missing for training, validation, and test, respectively.

\noindent \textbf{DAIC-WOZ} \cite{gratch2014distress} is a dataset composed of clinical interviews recorded in controlled settings to support the diagnosis of psychological distress conditions such as anxiety, depression, and post-traumatic stress disorder. It comprises 189 sessions with an average duration of 15.9 minutes, and it is highly imbalanced, containing 132 control and 57 depressed subjects. It is split into a training set with 107 samples, a validation set with 35 samples, and a test set with 47 samples. In addition to the transcript of the entire interview, COVAREP audio- \cite{degottex2014covarep} and OpenFace video-based \cite{morency2016openface} features are provided. Original recordings are not available for DAIC-WOZ for privacy reasons, and we used the original features provided by the authors. We used a total of 7 modalities: the acoustic COVAREP features, the 5 vocal tract resonance frequencies, the 68 3-dimensional facial landmarks, the facial action units \cite{friesen1978facial}, the gaze tracking, and the head pose landmarks. We did not use the text transcripts in our model. In our DAIC-WOZ experiment, we consider only the PHQ-8 binary labels. These binary labels are derived from the PHQ-8 score from each subject, with those who scored 10 or higher being labeled as depressed and the rest belonging to the control group.

\noindent \textbf{E-DAIC} \cite{ringeval2019avec} is the extended version of DAIC-WOZ \cite{gratch2014distress}. It comprises a total of 275 sessions with an average duration of 16.2 minutes, and it is severely imbalanced, containing 209 control and 66 depressed subjects. It is split into a training set with 163 samples, a validation set with 56 samples, and a test set with 56 samples. In addition to the interview transcript, multiple audio-visual features are provided based on both expert knowledge-based methods \cite{davis1980mfcc,eyben2016egemaps} and deep learning representations \cite{he2016resnet}. Similar to DAIC-WOZ, we only used the features provided by the authors. Concretely, we used a total of 6 modalities: the 13 mel frequency cepstral coefficients \cite{davis1980mfcc} alongside their first and second derivatives, the 88 eGeMAPS measures \cite{eyben2016egemaps} (covering information related to spectral, prosodic, and voice quality patterns), face embeddings extracted by a pre-trained ResNet-50 \cite{he2016resnet}, and the OpenFace features \cite{morency2016openface},  including gaze tracking, head pose landmarks and facial action units. We did not use the text transcripts in our model. Similar to DAIC-WOZ, we use the PHQ-8 binary labels as ground truth.

\begin{figure*}[!hbt]
   \begin{floatrow}
     \ffigbox{\includegraphics[scale = 0.3]{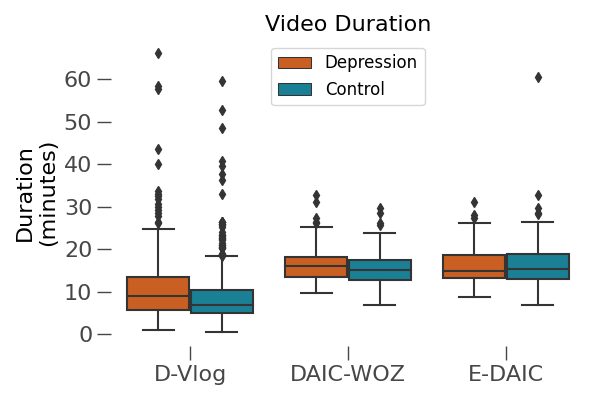}}
         {\caption{Distributions of video durations for each of our benchmarking datasets.}
        \label{fig:video-duration}}
     \ffigbox{\includegraphics[scale = 0.3]{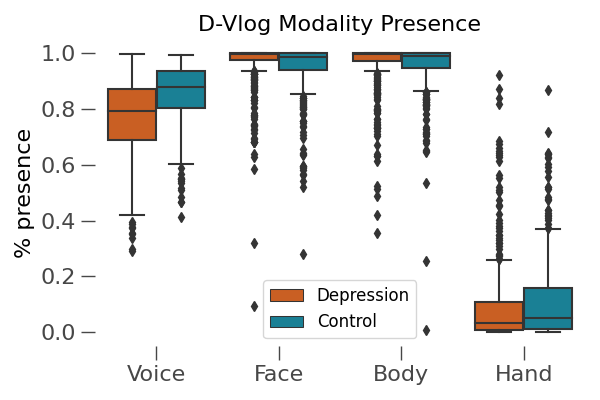}}
        {\caption{Presence distributions for each of our considered modalities in D-Vlog.}
         \label{fig:video-presence}}
   \end{floatrow}
\end{figure*}

\noindent \textbf{Discussion}.
We showcase the video durations for the datasets in Figure \ref{fig:video-duration}. DAIC-WOZ and E-DAIC videos have a longer overall duration (15.9 and 16.2 minutes, respectively) than D-Vlog and the video lengths vary considerably less. Notably, in D-Vlog, the depression group's videos are longer than those of the control group, with averages of 10.5 and 8.9 minutes, respectively. In Figure \ref{fig:video-presence}, we show the percentage of presence for each modality of interest in the D-Vlog dataset. Hands are the modality that is the least present in the vlogs. Even if the face and body are present in most of the videos, there are still many outliers in which the modalities are present less than 80\% of the time. As expected, the voice is also commonly present in the videos, but less in the depression group. The fact that in-the-wild vlogs may contain different transitions or sequences without the subject's presence in the video motivated us to incorporate the presence mask in our architecture.

\subsection{Evaluation}
Unlike other approaches, which either did not specify a formal evaluation protocol \cite{zheng2023two} or processed the features truncated to the average length of videos \cite{yoon2022dvlog,zhou2022tamfn}, we perform sequential evaluation across all temporal windows. For each non-overlapping window in a video $\{W_1, W_2, \dots W_n\} \in V_k$ taken in order (containing at least 50\% of modalities for D-Vlog), we perform depression classification and obtain a sequence of predictions: $\{y_1, y_2, \dots y_n\}$, with $y_l = f_{\hat{\theta}}(W_l)$. The final decision is performed by voting on the decisions across windows: $y = \mathrm{argmax}_{y_i} P(X = y_i)$. This evaluation protocol can be used for an early warning detection system \cite{coppersmith2018natural} for suicide prevention by taking the decision at an earlier window $n' < n$. For instance, this procedure can be used in live, continuous video streams on platforms such as Facebook or Twitch, in which cases of live suicide have been unfortunately registered in the past \cite{KAUSHIK2023100141}.

\subsection{Implementation Details}
Experiments were conducted on 2 GeForce RTX 3060 GPUs with 12GB memory. In all our experiments, we used the AdamW optimizer \cite{loshchilov2017decoupled} with a learning rate of 0.001 decaying through a cosine scheduler across 200 epochs, with a batch size of 8. On average, a training run took around 2 hours. In initial hyper-parameter search experiments, we found a suitable configuration for each model. Depending on the number of modalities used, our model's parameters range between 8.4M to 15M.  For D-Vlog, we used a context window size of 9 seconds and discarded window samples in which less than 50\% of the frames did not contain the subject's face. For DAIC-WOZ and E-DAIC, we did not discard any window, and used a context window of 6 seconds. For D-Vlog and E-DAIC, the transformer encoder was composed of 8 layers with 8 32-dimensional attention heads, while for DAIC-WOZ we used 8 layers with 4 64-dimensional attention heads. Similar to D-Vlog (see Subsection \ref{sec:non-verbal-cues}), for encoding the modalities in DAIC-WOZ and E-DAIC we defined a 2-layer transformer encoder when dealing with landmarks, and a linear projection when processing the other non-verbal features. For evaluation, we report mean and standard deviation across 5 training runs.

\subsection{Comparison Methods}
\label{sec:comparison}
\noindent \textbf{D-Vlog.} 
We compared our proposed method to several state-of-the-art approaches for depression detection from video. In contrast with previous works, which use only low-level facial and audio descriptors \cite{yoon2022dvlog,nguyenmultimodal,zheng2023two,zhou2022tamfn,zhou2023caiinet,tao2023depressive}, our model can also incorporate additional high-level facial and audio embeddings, gaze and blinking features, and hands and body landmarks. We also compare our method, which handles both local frame-level information and temporal video dynamics, to previous works that use global information from videos \cite{nguyenmultimodal,zhou2023caiinet,yoon2022dvlog}, and with previous time-aware methods \cite{zheng2023two,zhou2022tamfn,tao2023depressive}. We included two strong state-of-the-art methods \cite{zheng2023two,tao2023depressive}. Zheng et al. \cite{zheng2023two} proposed a temporal convolutional transformer that uses specialized medical and depression knowledge graphs. Tao et al. \cite{tao2023depressive}  proposed a spatio-temporal transformer architecture inspired by \cite{doersch2020crosstransformers,li2022shrinking}.
 
\noindent \textbf{DAIC-WOZ.} We compared our method to other works that approached the task from a non-verbal perspective, without considering the text transcripts. To show the capability of our proposed multi-modal architecture, we not only compared our method to those works that used hand-crafted features as our case \cite{williamson2016daicwoz,song2018daicwoz,wei2023daicwoz}, but also to those that explored deep-learning embedded representations \cite{ma2016daicwoz}. Furthermore, although Wei et al. \cite{wei2023daicwoz} introduced attention-based modules, most of these works were mainly based on CNNs combined with LSTMs \cite{ma2016daicwoz,song2018daicwoz,wei2023daicwoz}.

\noindent \textbf{E-DAIC.} Although prior works \cite{villatoro2021approximating,saggu2022depressnet,zheng2023two,ray2019multi} have been developed for this dataset, to the best of our knowledge, none of them explored the binary depression detection task from a non-verbal perspective, as all methods used the text transcript in their models. Consequently, we report the first results in this setting for E-DAIC.

Unlike our method, most previous works on D-Vlog only considered two modalities. Consequently, we introduce an additional strong baseline for general multi-modal classification for all three datasets. We used the Perceiver \cite{jaegle2021perceiver} based on an open-source implementation\footnote{\href{https://github.com/lucidrains/perceiver-pytorch}{https://github.com/lucidrains/perceiver-pytorch}, Accessed August 28th, 2023}, as a general multi-modal encoder. However, different from our approach, the Perceiver is based on iterative attention between modalities on a lower-dimensional hidden latent sequence, which can suffer from gradient explosion \cite{bengio1994learning,pascanu2013difficulty} and loss of information. We constructed the Perceiver to be similar to our model in terms of the number of parameters and hidden dimensions.

\section{Results}
\label{sec:results}
\begin{table}[hbt!]
    \begin{floatrow}
    \renewcommand\thetable{1}
    \ttabbox{%
   \caption{Results on the D-Vlog test set compared to existing approaches. Best results are highlighted in \textbf{bold} and second best results are \underline{underlined}.}
    }{%
     \begin{adjustbox}{width=0.50\textwidth}
    \begin{tabular}{lp{1.5cm}p{1.5cm}ccc}
         \textbf{Method} & \textbf{Original Split} & \textbf{Additional Modalities} & \textbf{F$_1$} & \textbf{Precision} & \textbf{Recall} \\ 
         \toprule
         {Yoon et al.} \cite{yoon2022dvlog} & \checkmark & \xmark & 0.64 & 0.65 & 0.66 \\
         {Nguyen et al.} \cite{nguyenmultimodal} & \checkmark & \xmark & 0.64 & 0.66 & 0.64 \\
         {Zheng et al.} \cite{zheng2023two} & \checkmark & \xmark & 0.65 & 0.65 & 0.65 \\
         {Zhou et al.} \cite{zhou2022tamfn} & \checkmark & \xmark & 0.66 & 0.66 & 0.67 \\
         {Zhou et al.} \cite{zhou2023caiinet} & \checkmark & \xmark & 0.67 & 0.67 & 0.67 \\ 
         {Tao et al.} \cite{tao2023depressive} & \checkmark & \xmark & \underline{0.75} & \textbf{0.73} & 0.78 \\ 
         \midrule
         {Perceiver} \cite{jaegle2021perceiver} & \checkmark & \xmark & 0.74{\scriptsize$\pm$0.01} & 0.67{\scriptsize$\pm$0.03} & \underline{0.84{\scriptsize$\pm$0.07}} \\
         {\textbf{Ours}} & \checkmark & \xmark & \textbf{0.76{\scriptsize$\pm$0.01}} & \underline{0.67{\scriptsize$\pm$0.02}} & \textbf{0.87{\scriptsize$\pm$0.02}} \\
         \midrule
         \midrule
         {Perceiver} \cite{jaegle2021perceiver} & \xmark & \xmark & 0.72{\scriptsize$\pm$0.01} & \underline{0.66{\scriptsize$\pm$0.03}} & 0.80{\scriptsize$\pm$0.03} \\
         {Perceiver} \cite{jaegle2021perceiver} & \xmark & \checkmark & 0.73{\scriptsize$\pm$0.01} & 0.68{\scriptsize$\pm$0.08} & 0.81{\scriptsize$\pm$0.15} \\
         {\textbf{Ours}} & \xmark & \xmark & \underline{0.73{\scriptsize$\pm$0.01}} & 0.62{\scriptsize$\pm$0.01} & \textbf{0.89{\scriptsize$\pm$0.05}} \\
         {\textbf{Ours}} & \xmark & \checkmark & \textbf{0.78{\scriptsize$\pm$0.01}$^\ddagger$} & \textbf{0.74{\scriptsize$\pm$0.05}} & \underline{0.84{\scriptsize$\pm$0.06}} \\
    \end{tabular}
    \end{adjustbox}
    \label{tab:dvlog-sota}
}
    \renewcommand\thetable{3}
    \ttabbox{%
    \caption{Results on the DAIC-WOZ test set. Best results are highlighted in \textbf{bold} and second best results are \underline{underlined}.}
    }{%
\begin{adjustbox}{width=0.40\textwidth}
    \begin{tabular}{lcccc}
         \textbf{Method} & \textbf{Modality} & \textbf{F\textsubscript{1}} & \textbf{Precision} & \textbf{Recall} \\ 
         \toprule
         S. Song et al. \cite{song2018daicwoz} & A & 0.46 & 0.32 & \underline{0.86} \\
         Ma et al. \cite{ma2016daicwoz} & A & 0.52 & 0.35 & \textbf{1.00} \\
         J. R. Williamson \cite{williamson2016daicwoz} & A & \underline{0.57} & - & - \\
         P-C. Wei et al. \cite{wei2023daicwoz} & A & \textbf{0.61} & \underline{0.56} & 0.66 \\
         \midrule
         {Perceiver} \cite{jaegle2021perceiver} & A & 0.33{\scriptsize$\pm$0.26} & 0.30{\scriptsize$\pm$0.28} & 0.45{\scriptsize$\pm$0.21} \\
         \textbf{Ours} & A & 0.49{\scriptsize$\pm$0.05} & \textbf{0.58{\scriptsize$\pm$0.05}} & 0.47{\scriptsize$\pm$0.05} \\
         \midrule
         \midrule
         S. Song et al. \cite{song2018daicwoz} & V & 0.50 & 0.60 & 0.43 \\
         J. R. Williamson \cite{williamson2016daicwoz} & V & 0.53 & - & - \\
         P-C. Wei et al. \cite{wei2023daicwoz} & V & 0.61 & 0.64 & 0.58 \\
         \midrule
         {Perceiver} \cite{jaegle2021perceiver} & V & \textbf{0.62{\scriptsize$\pm$0.05}} & \textbf{0.67{\scriptsize$\pm$0.08}} & \textbf{0.62{\scriptsize$\pm$0.05}} \\
         \textbf{Ours} & V & \underline{0.62{\scriptsize$\pm$0.06}} & \underline{0.66{\scriptsize$\pm$0.04}} & \underline{0.61{\scriptsize$\pm$0.06}} \\
         \midrule
         \midrule
         S. Song et al. \cite{song2018daicwoz} & AV & 0.50 & 0.60 & 0.43 \\
         P-C. Wei et al. \cite{wei2023daicwoz} & AV & 0.61 & \textbf{0.78} & 0.50 \\
         \midrule
         {Perceiver} \cite{jaegle2021perceiver} & AV & \underline{0.58{\scriptsize$\pm$0.13}} & \underline{0.75{\scriptsize$\pm$0.05}} & \underline{0.59{\scriptsize$\pm$0.11}} \\
         \textbf{Ours} & AV & \textbf{0.67{\scriptsize$\pm$0.05}} & 0.68{\scriptsize$\pm$0.04} & \textbf{0.66{\scriptsize$\pm$0.06}} \\
    \end{tabular}
    \end{adjustbox}
    \label{tab:daic-woz-sota}
    }

\end{floatrow}
\end{table}

\begin{table}[hbt!]
    \renewcommand\thetable{2}
    \begin{floatrow}

    \ttabbox{%
       \caption{Performance of our method on D-Vlog using different modality combinations, compared to the Perceiver. Best results are highlighted in \textbf{bold} and second best results are \underline{underlined}.}
    }{%
     \begin{adjustbox}{width=0.47\textwidth}
    \begin{tabular}{lcccc}

        \textbf{Method} & \textbf{Modality} & \textbf{F$_1$} & \textbf{Precision} & \textbf{Recall} \\ 
         \midrule
         \multirow{4}{*}{Perceiver \cite{jaegle2021perceiver}}  & {AV} & 0.69{\scriptsize$\pm$0.06} & 0.72{\scriptsize$\pm$0.07} & 0.69{\scriptsize$\pm$0.13} \\
          & {AV+EYES} & 0.73{\scriptsize$\pm$0.01} & 0.68{\scriptsize$\pm$0.08} & \underline{0.81{\scriptsize$\pm$0.15}} \\
          & {AV+LM} & 0.67{\scriptsize$\pm$0.13} & 0.70{\scriptsize$\pm$0.14} & 0.74{\scriptsize$\pm$0.28}\\
          & {AV+LM+EYES} & 0.64{\scriptsize$\pm$0.17} & \textbf{0.79{\scriptsize$\pm$0.07}} & 0.58{\scriptsize$\pm$0.22}  \\
         \midrule
         \multirow{4}{*}{\textbf{Ours}} & {AV} & 0.72{\scriptsize$\pm$0.05} & 0.73{\scriptsize$\pm$0.03} & 0.72{\scriptsize$\pm$0.11} \\
          & {AV+EYES} & \underline{0.75{\scriptsize$\pm$0.04}} & 0.72{\scriptsize$\pm$0.03} & 0.79{\scriptsize$\pm$0.12} \\
          & {AV+LM} & \underline{0.75{\scriptsize$\pm$0.03}} & 0.73{\scriptsize$\pm$0.05} & 0.77{\scriptsize$\pm$ 0.06}\\
          & {AV+LM+EYES} & \textbf{0.78{\scriptsize$\pm$0.01}} & \underline{0.74{\scriptsize$\pm$0.05}} & \textbf{0.84{\scriptsize$\pm$0.06}} \\
    \end{tabular}
    \end{adjustbox}
    \label{tab:dvlog-modalities}
}
    \renewcommand\thetable{4}
    \ttabbox{%
    \caption{Results on the E-DAIC test set. Best results are highlighted in \textbf{bold} and second best results are \underline{underlined}.}
    }{%
     \begin{adjustbox}{width=0.43\textwidth}
    \begin{tabular}{lcccc}
        \textbf{Method} & \textbf{Modality} & \textbf{F$_1$} & \textbf{Precision} & \textbf{Recall} \\
         \midrule
         \multirow{3}{*}{Perceiver \cite{jaegle2021perceiver}}  & A & 0.36{\scriptsize$\pm$0.17} & \underline{0.66{\scriptsize$\pm$0.12}} & 0.41{\scriptsize$\pm$0.09} \\
          & V & 0.39{\scriptsize$\pm$0.22} & 0.50{\scriptsize$\pm$0.26} & 0.48{\scriptsize$\pm$0.18} \\
         & AV & \underline{0.53{\scriptsize$\pm$0.09}} & 0.54{\scriptsize$\pm$0.05} & \underline{0.54{\scriptsize$\pm$0.11}} \\
         \midrule
         \multirow{3}{*}{\textbf{Ours}} & A & 0.51{\scriptsize$\pm$0.11} & \textbf{0.71{\scriptsize$\pm$0.05}} & 0.52{\scriptsize$\pm$0.08} \\
          & V & 0.50{\scriptsize$\pm$0.08} & 0.58{\scriptsize$\pm$0.05} & 0.49{\scriptsize$\pm$0.07} \\
         & AV & \textbf{0.56{\scriptsize$\pm$0.12}} & 0.59{\scriptsize$\pm$0.07} &\textbf{ 0.58{\scriptsize$\pm$0.13}} \\
    \end{tabular}
    \end{adjustbox}
    \label{tab:e-daic-sota}
    }

\end{floatrow}
\end{table}

\noindent \textbf{D-Vlog.} 
In Table \ref{tab:dvlog-sota}, we present the results of our method, compared to previous state-of-the-art approaches. As presented in Subsection \ref{sec:datasets}, since the release of the D-Vlog dataset, some YouTube videos became unavailable, leading to a reduction of the test split. To provide a fair comparison, we evaluate our model on both the original test split, and also on the test split of currently available videos from which we extracted additional modalities. Our model achieves a 0.76 F$_1$ using the original low-level facial and audio features provided by D-Vlog authors from the original test split. However, when we evaluate our model using the original features on the reduced test split, the performance of the model is lower, with only a 0.73 F$_1$, pointing to the fact that the samples from the reduced test set are more challenging to classify than the original test split. Using the additional modalities (audio-visual, landmarks and eyes) inspired by psychological research, our model obtains the best performance of 0.78 F$_1$, surpassing the previous state-of-the-art method proposed by Tao et al. \cite{tao2023depressive}. Moreover, compared to the Perceiver, a state-of-the-art method for multi-modal classification, our model achieves a considerable improvement of 5\% F$_1$.

We provide the results of our method using different modalities in Table \ref{tab:dvlog-modalities}. Even if the Perceiver model is a strong baseline for multi-modal information processing, our proposed method surpasses it in all settings. Adding \textbf{LM} (face, body and hands landmarks) and \textbf{EYES} features (gaze and blinking) separately offers a slight improvement over the \textbf{AV} setting using only face and audio embeddings. Our model obtains the best performance with all modalities \textbf{AV+LM+EYES}, achieving an F$_1$ of 0.78. Results show that different non-verbal manifestations of depression are needed for accurate detection from video data. Our best-performing model uses emotional-informed face embeddings from EmoNet, audio embeddings, spatial information from the face, body and hand landmarks, and gaze and blinking patterns. 





\noindent \textbf{DAIC-WOZ.} Even though the DAIC-WOZ dataset was extensively used for depression detection, most previous works incorporate text-based features extracted from the video transcripts, with state-of-the-art approaches achieving performances of around 0.96 F$_1$ \cite{zheng2023two}. However, this high performance is not due to a deep understanding of depression, but is rather due to spurious correlations between the text transcript and the depression label, as the dataset contains recordings of clinical interviews with predefined questions from PHQ-8. For this reason, we are using only non-verbal cues, and discard the text transcript, to provide a more realistic performance measurement for depression detection. 
The methods we considered in Table \ref{tab:daic-woz-sota} do not incorporate textual features for depression modeling.  Our model is competitive with other methods in video-only settings, but we achieve the best performance of 0.67 F$_1$ using both modalities. Compared to the Perceiver baseline, our method performs better in audio and audio-visual settings.




\noindent \textbf{E-DAIC.} Similar to DAIC-WOZ, the current state of the art for binary classification explored text-based features extracted from the transcripts, a considerably easier formulation than only using non-verbal cues, achieving high performance \cite{zheng2023two}. In Table \ref{tab:e-daic-sota}, we report the first results addressing the task from a non-verbal perspective on binary classification for E-DAIC. Our model obtains a substantial improvement when combining both audio and visual modalities, and outperforms the Perceiver in all scenarios, showing that our architecture is highly efficient at integrating multiple modalities.





\begin{figure}[hbt!]
    \centering
    \includegraphics[width=0.8\textwidth]{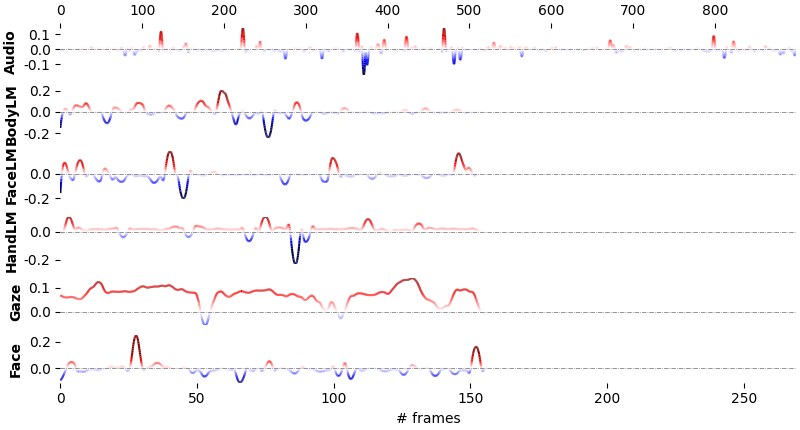}
    \caption{Attribution scores per each modality across frames obtained with Integrated Gradients \cite{sundararajan2017axiomatic} on a selected window from a subject suffering from depression from D-Vlog. Higher values correspond to a strong attribution towards a positive prediction.}
    \label{fig:explainability}
\end{figure}

\noindent \textbf{Explainability.} Any model developed to work with data from humans, especially in clinical scenarios, needs to be interpretable and explainable, to increase trust and transparency, ethical and legal compliance, and to allow researchers and engineers to debug and perform error analysis. Consequently, we present a potential way of explaining multi-modal depression detection models from video using Integrated Gradients \cite{sundararajan2017axiomatic}. In Figure \ref{fig:explainability}, we show attribution scores for 6 modalities in a selected window from a video from D-Vlog, with higher values corresponding to a strong attribution towards a positive prediction. We show the attribution values across time, for the video frames (bottom) and the audio frames (top), sampled at a higher frequency. In this particular context window, movement through the body and hand landmarks and gaze are more indicative of the mental state of the subject, compared to audio or facial expressions. From around frame 150, the person is not visible in the video (while audio is still present), and consequently, the attribution scores for each missing modality are 0. This shows that our model can properly handle missing modalities. Our analysis provides the first steps in automatic interpretation and understanding of how different non-verbal cues contribute to the manifestations of depression.

\noindent \textbf{Limitations and Ethical Considerations}. 
Our approach has shown promising results in detecting depression through non-verbal behavior, but it is important to note that it should not be used as a means of clinical diagnosis, which should only be done by a mental health professional. However, our method can be useful for initial depression screening by identifying psychomotor manifestations. In our experiments, we did not include any demographic information about the subjects; the results of our approach may differ when applied to other demographic groups \cite{oureshi2021gender,bailey2021gender}. Lastly, the data we used in our experiments were anonymized, and we made no attempt to contact the subjects.

\section{Conclusion}
\label{sec:conclusion}
In this work, we presented a simple and flexible multi-modal transformer architecture capable of detecting depression from multiple non-verbal cues from video. Following psychological studies \cite{bennabi2013psychomotor,renneberg2005facial,fiquer2013talking,yamamoto2020using}, we explored additional high-level modalities and showed a markedly improvement in depression detection from in-the-wild videos. We obtained state-of-the-art results in three key benchmarking datasets for depression detection, both in-the-wild vlogs \cite{yoon2022dvlog} and in recorded clinical interviews \cite{gratch2014distress,ringeval2019avec}, surpassing previous works \cite{yoon2022dvlog,nguyenmultimodal,zheng2023two,zhou2022tamfn,zhou2023caiinet,tao2023depressive,song2018daicwoz,ma2016daicwoz,wei2023daicwoz,williamson2016daicwoz} by a considerable margin. Finally, we showed that our model is interpretable, and it can provide importance scores of each modality of a particular subject across time, making our method a viable solution for depression estimation, potentially in early-warning detection and suicide prevention in platforms with continuous video streams.

\begin{sloppypar}
\noindent \textbf{Acknowledgements.} The work of David Gimeno-Gómez and Carlos-D. Martínez-Hinarejos was partially supported by Grant CIACIF/2021/295 funded by Generalitat Valenciana and by Grant PID2021-124719OB-I00 under project LLEER funded by MCIN/AEI/10.13039/501100011033/ and by ERDF, EU A way of making Europe. The work of Paolo Rosso was in the framework of the PID2021-124361OB-C31 research project funded by MCIN/AEI/10.13039/501100011033 and by ERDF, EU A way of making Europe.
\end{sloppypar}

%
%
%
\bibliographystyle{splncs04}
\bibliography{refs}

\end{document}